\documentclass[final]{cvpr}

\usepackage{times}
\usepackage{epsfig}
\usepackage{graphicx}
\usepackage{amsmath}
\usepackage{amssymb}
\usepackage{subcaption}

\usepackage{bbding}
\usepackage{pifont}
\usepackage{color}
\usepackage{bm}

\usepackage[pagebackref=true,breaklinks=true,colorlinks,bookmarks=false]{hyperref}

\newcommand{\cmark}{\ding{51}}%
\newcommand{\xmark}{\ding{55}}%

\makeatletter
\newcommand{\printfnsymbol}[1]{%
	\textsuperscript{\@fnsymbol{#1}}%
}
\makeatother

\def\cvprPaperID{886} 
\def\confYear{CVPR 2021}

\begin{document}

\title{MASA-SR: Matching Acceleration and Spatial Adaptation for\\ Reference-Based Image Super-Resolution}

\author{
	Liying Lu\textsuperscript{1}\thanks{Equal contribution} \quad Wenbo Li\textsuperscript{1}\printfnsymbol{1} \quad Xin Tao\textsuperscript{2} \quad  Jiangbo Lu\textsuperscript{3} \quad Jiaya Jia\textsuperscript{1,3}\\
	$^1$ The Chinese University of Hong Kong \qquad
	$^2$ Kuaishou \qquad
	$^3$ SmartMore\\
	{\tt\small \{lylu,wenboli,leojia\}@cse.cuhk.edu.hk, jiangsutx@gmail.com, jiangbo@smartmore.com} 
}

\maketitle

\begin{abstract}
	Reference-based image super-resolution (RefSR) has shown promising success in recovering high-frequency details by utilizing an external reference image (Ref). In this task, texture details are transferred from the Ref image to the low-resolution (LR) image according to their point- or patch-wise correspondence. Therefore, high-quality correspondence matching is critical. It is also desired to be computationally efficient. 
	Besides, existing RefSR methods tend to ignore the potential large disparity in distributions between the LR and Ref images, which hurts the effectiveness of the information utilization. In this paper, we propose the MASA network for RefSR, where two novel modules are designed to address these problems. 
	The proposed Match~\&~Extraction Module significantly reduces the computational cost by a coarse-to-fine correspondence matching scheme. The Spatial Adaptation Module learns the difference of distribution between the LR and Ref images, and remaps the distribution of Ref features to that of LR features in a spatially adaptive way. This scheme makes the network robust to handle different reference images. Extensive quantitative and qualitative experiments validate the effectiveness of our proposed model. Codes are available at \href{https://github.com/dvlab-research/MASA-SR}{https://github.com/dvlab-research/MASA-SR}.
	
\end{abstract}

\vspace{-0.2in}
\section{Introduction}

\begin{figure}[t]
	\begin{center}
		\includegraphics[width=1.0\linewidth]{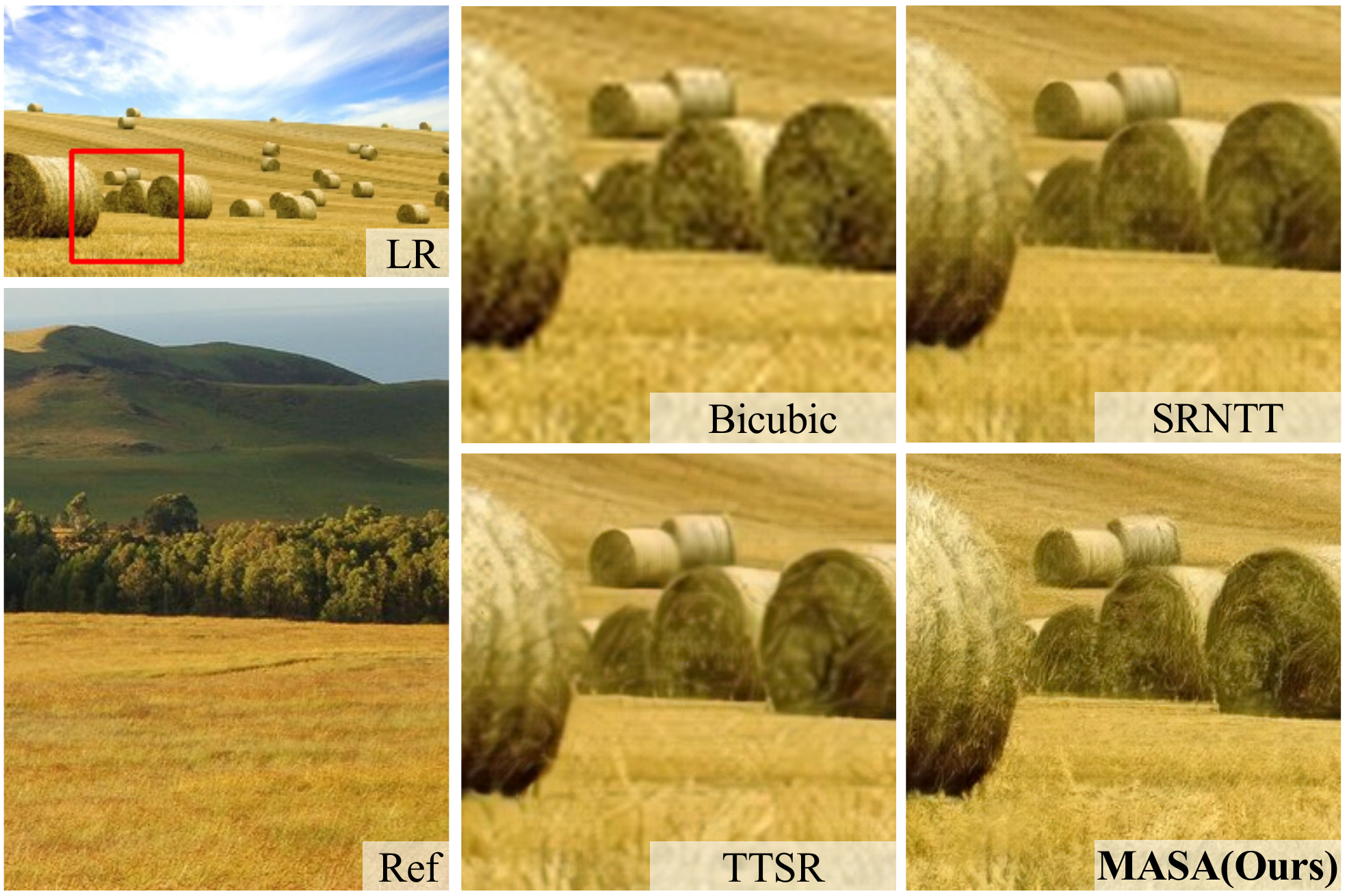}
	\end{center}
	\caption{Visual comparison of $\times 4$ SR results. Our MASA method generates more appealing texture details than the two leading RefSR methods, \ie, SRNTT~\cite{srntt} and TTSR~\cite{ttsr}.} 
	\label{fig:teaser}
\end{figure}

Single image super-resolution (SISR) is a fundamental computer vision task that aims to restore a high-resolution image (HR) with high-frequency details from its low-resolution counterpart (LR). Progress of SISR in recent years is based on deep convolutional neural networks (CNN)~\cite{srcnn,vdsr,drcn,lapsr,edsr,rcan}. Nevertheless, the ill-posed nature of SISR problems makes it still challenging to recover high-quality details.
 
In this paper, we explore reference-based super-resolution (RefSR), which utilizes an external reference image (Ref) to help super-resolve the LR image. Reference images usually contain similar content and texture with the LR image. They can be acquired from web image search or captured from different viewpoints. Transferring fine details to the LR image can overcome the limitation of SISR and has demonstrated promising performance in recent work of~\cite{crossnet,shim2020robust,zheng2017learning,srntt,ttsr}.

Previous methods aimed at designing various ways to handle two critical issues in this task: \textit{a) Correspond useful content in Ref images with LR images. b) Transfer features from Ref images to facilitate HR image reconstruction}.
To address the first issue, methods perform spatial alignment between the Ref and LR images~\cite{crossnet,shim2020robust} using optical flow or deformable convolutions~\cite{deformable,deformablev2}. These alignment-based methods face challenges in, \eg, finding long-distance correspondence. Other methods follow patch matching~\cite{zheng2017learning,srntt,ttsr} in the feature space. State-of-the-art methods generally perform dense patch matching, leading to very high computational cost and large memory usage.
 
For the second issue, our finding is that even if LR and Ref images share similar content, color and luminance may differ. Previous methods directly concatenate the LR features with the Ref ones and fuse them in convolution layers, which is not optimal. 

To address the above problems, we propose a RefSR method called MASA-SR, which improves patch matching and transfer. The design of MASA has several advantages. 
First, the proposed Match~\&~Extraction Module (MEM) performs correspondence matching in a coarse-to-fine manner, which largely reduces the computational cost while maintaining the matching quality. By leveraging the local coherence property of natural images, for each patch in the LR feature maps, we shrink its search space from the whole Ref feature map to a specific Ref block.

Second, the Spatial Adaptation Module is effective in handling the situations where there exists large disparity in color or luminance distribution between the LR and Ref images. It learns to remap the distribution of the Ref features to LR ones in a spatially adaptive way. Useful information in the Ref features thus can be transferred and utilized more effectively.

To the best of our knowledge, our model achieves state-of-the-art performance for the RefSR task. Our contributions are as follows.
\begin{itemize}
	\item The proposed Match~\&~Extraction Module significantly reduces the computational cost of correspondence matching in the deep feature space. Our results show that  a {\it two-orders-of-magnitude} reduction measured in FLOPS is achieved.
	\item The proposed Spatial Adaptation Module is robust to Ref images with different color and luminance distributions. It enables the network to better utilize useful information extracted from Ref images.
\end{itemize}

\section{Related Work}

\subsection{Single Image Super-Resolution}

Effort has been made to improve performance of single image super resolution in recent years. In particular, deep learning based SISR methods achieved impressive success. Dong \etal~\cite{srcnn} proposed the seminal CNN-based SISR model that consists of three convolution layers. Later, a variety of effective networks~\cite{vdsr,drcn,lapsr,edsr,rcan,lapar,drrn} were proposed for SR. With the help of residual learning, Kim \etal proposed VDSR~\cite{vdsr} and DRCN~\cite{drcn} with deeper architectures and improved accuracy. Lai \etal~\cite{lapsr} proposed LapSRN, which progressively reconstructs multi-scale results in a pyramid framework. Lim \etal~\cite{edsr} removed batch normalization layers in residual networks and further expanded the model size to improve SR performance. Zhang \etal~\cite{rcan} built a very deep network with residual in residual structure. Channel attention was introduced to model the inter-dependency across different channels. 

Apart from MSE minimizing based methods, perception-driven ones received much attention. The perceptual loss~\cite{johnson2016perceptual} was introduced into SR tasks to enhance visual quality by minimizing errors on high-level features. Ledig \etal~\cite{srgan} proposed SRGAN, which was trained with an adversarial loss, generating photo-realistic images with natural details. To produce more perceptually satisfying results, ESRGAN~\cite{esrgan} further improves SRGAN by introducing a relativistic adversarial loss. Different from the adversarial loss, the contextual loss~\cite{contextual,contextual2,idmrf} was proposed to maintain natural statistics in generated images by measuring the feature distribution.

\subsection{Reference-Based Super-Resolution}

Compared with SISR, which only takes as input a low-resolution image, RefSR uses an additional reference image to upsample the LR input. The reference image generally has similar content with the LR image, capable to provide high-frequency details. 
Recent work mostly adopts CNN-based frameworks. One branch of RefSR performs spatial alignment between the Ref and LR images. CrossNet \cite{crossnet} estimated flow between the Ref and LR images at multi-scales and warped the Ref features according to the flow. However, the flow was obtained by a pre-trained network, leading to heavy computation and inaccurate estimation. Shim \etal~\cite{shim2020robust} further proposed to align and extract Ref features by leveraging deformable convolutions~\cite{deformable,deformablev2}. Nevertheless, these alignment-based methods are limited in finding long-distance correspondence.

Another branch follows the idea of patch matching~\cite{patchmatch}. Zheng \etal~\cite{zheng2017learning} trained two networks to learn feature correspondence and patch synthesis respectively. SRNTT~\cite{srntt} conducted multi-level patch matching between Ref and LR features extracted from the pre-trained VGG~\cite{vgg}, and fused the swapped Ref features together with the LR features to generate the SR result. TTSR~\cite{ttsr} further introduced the transformer architecture into the RefSR task and stacked the transformer in a cross-scale way to fuse multi-level information. The hard attention and soft attention in the transformer help transfer texture features from the Ref image more precisely. However, the patch matching method of SRNTT and TTSR is of high computation cost. They also leveraged VGG as the feature extractor that is heavy and requires pre-training.

\begin{figure*}[h]
	\begin{center}
		\includegraphics[width=1.0\linewidth]{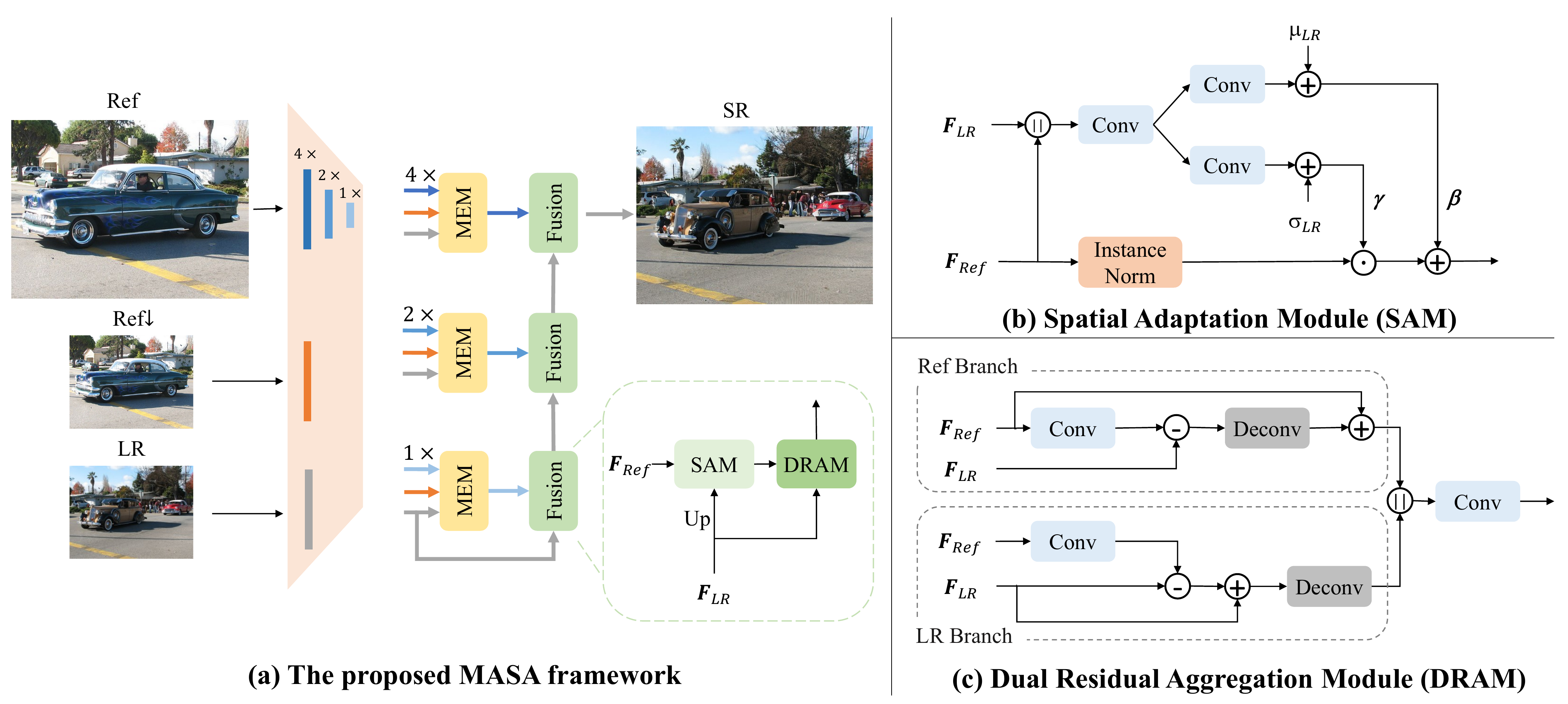}
	\end{center}
	\vspace{-0.18in}
	\caption{(a) Framework of the proposed MASA-SR, which consists of an encoder, several Match~\&~Extraction Modules~(MEM), Spatial Adaptation Modules~(SAM) and Dual Residual Aggregation Modules~(DRAM). (b) Structure of the Spatial Adaptation Modules~(SAM), which is used to remap the distributions of the Ref features to that of the LR features. (c) Structure of the Dual Residual Aggregation Modules~(DRAM), which is effective in feature fusion. }
	\label{fig:network}
\vspace{-0.12in}
\end{figure*}

\section{MASA-SR Method}

As shown in Fig.~\ref{fig:network}(a), our framework mainly consists of three parts: the encoder, \textit{Matching~\&~Extraction Modules~(MEM)}, and fusion modules that contain \textit{Spatial Adaptation Modules~(SAM)} and \textit{Dual Residual Aggregation Modules~(DRAM)}. LR, Ref$\downarrow$ and Ref denote the low-resolution image, the $\times4$ bicubic-downsampled reference image and the reference image, respectively. Unlike previous methods~\cite{srntt,ttsr} that use the pre-trained VGG as the feature extractor, our encoder is trained along with other parts of the network from scratch.

The encoder consists of three building blocks -- the second and third blocks halve the size of the feature maps with stride 2. After passing the Ref image into the encoder, three Ref features with different scales are obtained as ${\bm F}_{Ref}^{s}$, where $s=1,2,4$.
The LR image and the Ref$\downarrow$ image only go through the first block of the encoder, producing ${\bm F}_{LR}$ and ${\bm F}_{Ref\downarrow}$.

Afterwards, $\{ {\bm F}_{LR}, {\bm F}_{Ref\downarrow}, {\bm F}_{Ref}^{s} \}$ are fed into MEM to perform coarse-to-fine correspondence matching and feature extraction as shown in Fig.~\ref{fig:MTM}. Though there are three MEMs in Fig.~\ref{fig:network}(a), the matching steps are only performed once between ${\bm F}_{LR}$ and ${\bm F}_{Ref\downarrow}$. The feature extraction stage is performed three times, each for one Ref feature ${\bm F}_{Ref}^{s}$ of scale $s$. To generate the final SR output, the LR features and output features from MEM are fused through the fusion module, where the proposed SAM is used to align the statistics of Ref features to those of LR ones. The proposed DRAM is used to enhance high-frequency details.

MEM, SAM and DRAM are explained in Sections~\ref{sec:mem}, \ref{sec:pa} and \ref{sec:dram}, respectively. The loss functions used to train the network are introduced in Section~\ref{sec:loss}.
 
\subsection{Matching~\&~Extraction Module~(MEM)}
\label{sec:mem}

It is known that in a local region of a natural image, neighboring pixels are likely to come from common objects and share similar color statistics. Previous research on natural image priors also indicates that neighboring patches in one image are likely to find their correspondence spatially coherent with each other. 

\begin{figure}[t]
	\begin{center}
		\includegraphics[width=1.0\linewidth]{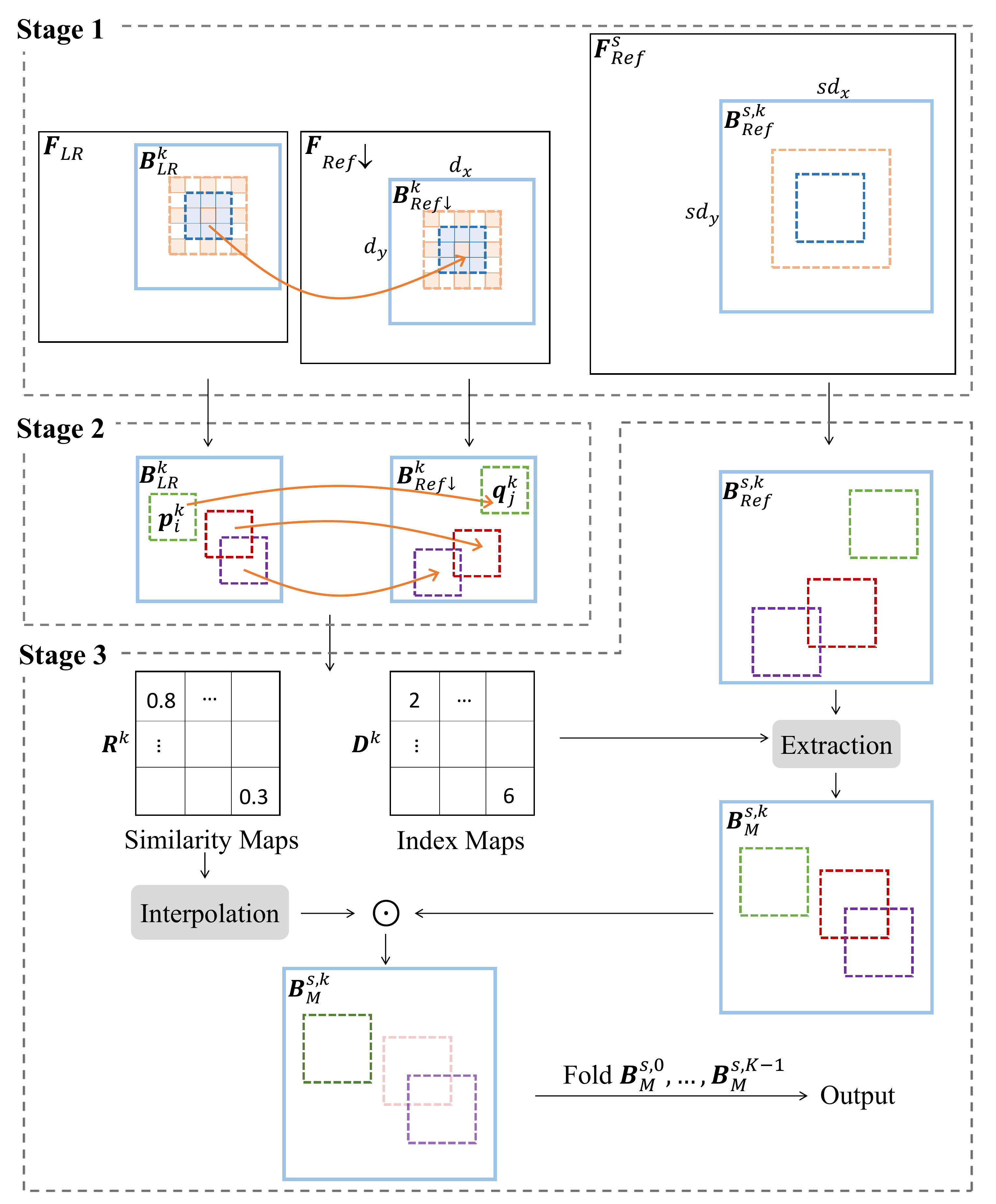}
	\end{center}
	\vspace{-0.18in}
	\caption{Pipeline of the Match~\&~Extraction Module~(MEM). In {\bf Stage 1}, each LR block finds the most relevant Ref$\downarrow$ block. In {\bf Stage 2}, dense patch matching is performed in each (LR block, Ref$\downarrow$ block) pair. In {\bf Stage 3}, Ref features are extracted according to the similarity and index maps produced in the second stage. All the blocks are denoted by solid squares in light blue and patches are denoted by dotted squares in different colors. }
	\label{fig:MTM}
\vspace{-0.12in}
\end{figure}

This motivates us to propose a coarse-to-fine matching scheme, {\ie, coarse block matching and fine patch matching. Note that `block' and `patch' are two different concepts in our method, and the size of block is larger than patch~($3\times 3$ in our experiments).} As shown in Fig.~\ref{fig:MTM}, we first find correspondences in the feature space only for blocks. Specifically, we unfold the LR feature into non-overlapping blocks. Each LR block will find its most relevant Ref$\downarrow$ block. By doing so, the computational cost of matching is reduced significantly compared with previous methods~\cite{srntt,ttsr}. To achieve enough precision, we further perform dense patch matching within each (LR block, Ref$\downarrow$ block) pair. 
In the last stage, we extract useful Ref features according to the obtained correspondence information.

\vspace{4pt}
\noindent\textbf{Stage 1: Coarse matching. } In this stage, the LR feature ${\bm F}_{LR}$ is unfolded into $K$ non-overlapping blocks: $\{ {\bm B}_{LR}^{0}, ..., {\bm B}_{LR}^{K-1} \}$. For each LR block ${\bm B}_{LR}^{k}$, we find its most relevant Ref$\downarrow$ block ${\bm B}_{Ref\downarrow}^{k}$. 

We first take the center patch of ${\bm B}_{LR}^{k}$ to compute the cosine similarity with each patch of ${\bm F}_{Ref\downarrow}$ as
\begin{align}
	r_{c,j}^{k} = \left \langle \frac{{\bm p}^{k}_{c}}{\left \| {\bm p}^{k}_{c} \right \|}, \frac{{\bm q}_{j}}{\left \| {\bm q}_{j} \right \|} \right \rangle \;,
\end{align}
where ${\bm p}^{k}_{c}$ is the center patch of ${\bm B}_{LR}^{k}$, ${\bm q}_{j}$ is the $j$-th patch of ${\bm F}_{Ref\downarrow}$, and $r_{c,j}^{k}$ is their similarity score. According to the similarity scores, we find the most similar patch for ${\bm p}^{k}_{c}$ in ${\bm F}_{Ref\downarrow}$. We then crop the block of size $d_{x}\times d_{y}$ centered around this similar patch, denoted as ${\bm B}_{Ref\downarrow}^{k}$. According to the local coherence property, for all patches in ${\bm B}_{LR}^{k}$, their most similar patches are likely to reside in this ${\bm B}_{Ref\downarrow}^{k}$. 
On the other hand, we also crop the corresponding $sd_{x}\times sd_{y}$ block from ${\bm F}_{Ref}^{s}$, denoted by ${\bm B}_{Ref}^{s,k}$, which will be used in the feature extraction stage.

Note that the center patch may not be representative enough to cover full content of the LR block if the size of the LR block is much larger than that of its center patch. This may mislead us to find the irrelevant Ref$\downarrow$ block.
To address it, we use center patches with different dilation rates to compute the similarity. The details are shown in Stage 1 of Fig.~\ref{fig:MTM}, where the dotted blue patch denotes the case of $dilation=1$ and the dotted orange patch denotes the case of $dilation=2$.
Then the similarity score is computed as the sum of results of different dilations.

After this stage, for each LR block, we obtain its most relevant Ref$\downarrow$ block and the corresponding Ref block, forming triples of (${\bm B}_{LR}^{k}$, ${\bm B}_{Ref\downarrow}^{k}$, ${\bm B}_{Ref}^{s,k}$).
We limit the search space of ${\bm B}_{LR}^{k}$ to ${\bm B}_{Ref\downarrow}^{k}$ in the fine matching stage.

\vspace{4pt}
\noindent\textbf{Stage 2: Fine matching. } In this stage, dense patch matching is performed between each LR block and its corresponding Ref$\downarrow$ block independently. A set of index maps $\{ {\bm D}^{0},..., {\bm D}^{K-1} \}$ and similarity maps $\{ {\bm R}^{0},..., {\bm R}^{K-1} \}$ are obtained.

More precisely, taking the $k$-th pair (${\bm B}_{LR}^{k}$, ${\bm B}_{Ref\downarrow}^{k}$) for example, we compute the similarity score between each patch of ${\bm B}_{LR}^{k}$ and each patch of ${\bm B}_{Ref\downarrow}^{k}$ as
\begin{align}
	r_{i,j}^{k} = \left \langle \frac{{\bm p}_{i}^{k}}{\left \| {\bm p}_{i}^{k} \right \|}, \frac{{\bm q}_{j}^{k}}{\left \| {\bm q}_{j}^{k} \right \|} \right \rangle,
\end{align}
where ${\bm p}_{i}^{k}$ is the $i$-th patch of ${\bm B}_{LR}^{k}$, ${\bm q}_{j}^{k}$ is the $j$-th patch of ${\bm B}_{Ref\downarrow}^{k}$, and $r_{i,j}^{k}$ is their similarity score. 
Then the $i$-th element of ${\bm D}^{k}$ is calculated as
\begin{align}
{\bm D}_{i}^{k} = \mathop{\arg\max}_{j} \ \ r_{i,j}^{k} \;.
\end{align}
The $i$-th element of ${\bm R}^{k}$ is the highest similarity score related to the $i$-th patch of ${\bm B}_{LR}^{k}$ as
\begin{align}
{\bm R}_{i}^{k} = \mathop{\max}_{j} \ \ r_{i,j}^{k} \;.
\end{align}

\noindent\textbf{Stage 3: Feature extraction. } In this stage, we first extract patches from ${\bm B}_{Ref}^{s,k}$ according to the index map ${\bm D}^{k}$, and form a new feature map ${\bm B}_{M}^{s,k}$. Specifically, We crop the ${\bm D}_{i}^{k}$-th patch of ${\bm B}_{Ref}^{s,k}$ as the $i$-th patch of ${\bm B}_{M}^{s,k}$.
Moreover, since Ref features with higher similarity scores are more useful, we multiply ${\bm B}_{M}^{s,k}$ with the corresponding similarity score map ${\bm R}^{k}$ to get the weighted feature block as
\begin{align}
{\bm B}_{M}^{s,k} := {\bm B}_{M}^{s,k} \odot ({\bm R}^{k})\uparrow \;,
\end{align}
where ()$\uparrow$ and $\odot$ denote bilinear interpolation and element-wise multiplication. 

The final ouput of MEM is obtained by folding $\{ {\bm B}_{M}^{s,0}, ..., {\bm B}_{M}^{s,K-1} \}$ together, which is the reverse operation of the unfolding operation in Stage 1.

\vspace{4pt}
\noindent\textbf{Analysis. } For an LR image with $m$ pixels and a Ref$\downarrow$ image with $n$ pixels, computational complexity of matching in previous methods is generally $\rm O$$(mn)$. While in the MEM, suppose each Ref$\downarrow$ block has $n'$ pixels, computational complexity is reduced to $\rm O$$(Kn+mn')$. Since $K$ is much smaller than $m$ and $n'$ is also several hundred times smaller than $n$, the computational cost is reduced significantly through this coarse-to-fine matching scheme.

\subsection{Spatial Adaptation Module~(SAM)}
\label{sec:pa}

In many situations, the LR and the Ref images may have similar content and texture. But color and luminance distributions diverge. Thus the distribution of extracted Ref features may not be consistent with that of the LR features. Therefore, simply concatenating the Ref and LR features together and feeding them into the following convolution layers is not optimal.
Inspired by~\cite{analogies,spade}, we propose the Spatial Adaptation Module~(SAM) to remap the distribution of the extracted Ref features to that of the LR features.

We illustrate the structure of SAM in Fig.~\ref{fig:network}b. The LR feature and extracted Ref feature are first concatenated before feeding into convolution layers to produce two parameters $\bm \beta$ and $\bm \gamma$, which are with the same size as the LR feature. Then instance normalization~\cite{instancenorm} is applied to the Ref feature as
\begin{align}
{\bm F}_{Ref}^{c} \longleftarrow \frac{{\bm F}_{Ref}^{c}-{\bm \mu}_{Ref}^{c}}{{\bm \sigma}_{Ref}^{c}} \;,
\end{align}
where ${\bm \mu}_{Ref}^{c}$ and ${\bm \sigma}_{Ref}^{c}$ are the mean and standard deviation of ${\bm F}_{Ref}$ in channel $c$ as 
\begin{align}
{\bm \mu}_{Ref}^{c} = \frac{1}{HW} \sum_{y,x} {\bm F}_{Ref}^{c,y,x} \;, \label{eq:mu}
\end{align}
\vspace{-0.25in}
\begin{align}
{\bm \sigma}_{Ref}^{c} = \sqrt{\frac{1}{HW} \sum_{y,x} \big ( {\bm F}_{Ref}^{c,y,x}-{\bm \mu}_{Ref}^{c}\big )^{2}} \;. \label{eq:sigma}
\end{align}
$H$ and $W$ are the height and width of ${\bm F}_{Ref}$ .We then update ${\bm \beta}$ and ${\bm \gamma}$ with the mean and standard deviation of the LR feature of
\begin{align}
{\bm \beta} \longleftarrow {\bm \beta} + {\bm \mu}_{LR} \;, \\
{\bm \gamma} \longleftarrow {\bm \gamma} + \bm {\sigma}_{LR} \;,
\end{align}
where ${\bm \mu}_{LR}$ and ${\bm \sigma}_{LR}$ are computed in a similar way as Eqs.~(\ref{eq:mu}) and (\ref{eq:sigma}).
Finally, ${\bm \gamma}$ and ${\bm \beta}$ are multiplied and added to the normalized Ref feature in an element-wise manner as
\begin{align}
{\bm F}_{Ref} \longleftarrow {\bm F}_{Ref} \cdot {\bm \gamma} + {\bm \beta} \;.
\end{align} 

Since the difference between the Ref features and LR features varies with respect to the spatial location, while the statistics ${\bm \mu}_{LR}$, ${\bm \sigma}_{LR}$, ${\bm \mu}_{Ref}$ and ${\bm \sigma}_{Ref}$ are of size $C\times 1\times 1$, we use learnable convolutions to predict two spatial-wise adaptation parameters ${\bm \beta}$ and ${\bm \gamma}$. Unlike \cite{spade} that only uses the segmentation maps to produce two parameters, the convolutions in SAM takes as the input both Ref and LR features to learn their difference. Besides, after obtaining ${\bm \beta}$ and ${\bm \gamma}$ from the convolutions, we add them with the mean and standard deviation of the LR features.

\subsection{Dual Residual Aggregation Module (DRAM)}
\label{sec:dram}

After spatial adaptation, the transferred Ref features are fused with the LR features using our proposed Dual Residual Aggregation Module (DRAM) as shown in Fig.~\ref{fig:network}(c). DRAM consists of two branches, \ie, the LR branch and the Ref branch.

The Ref branch aims to refine the high-frequency details of the Ref features. It first downsamples the Ref feature  ${\bm F}_{Ref}$ by a convolution layer with stride 2, and the residual ${\bm Res}_{Ref}$ between the downsampled Ref feature and the LR feature ${\bm F}_{LR}$ is then upsampled by a transposed convolution layer as
\begin{equation}
\left\{
	\begin{array}{lr}
	{\bm Res}_{Ref} = Conv({\bm F}_{Ref}) - {\bm F}_{LR} \;, \\
	{\bm F}_{Ref}^{'} = {\bm F}_{Ref} + Deconv({\bm Res}_{Ref}) \;.
	\end{array}
\right.
\end{equation} 
Similarly, the high-frequency details of the LR features are refined as
\begin{equation}
	\left\{
	\begin{array}{lr}
	{\bm Res}_{LR} = {\bm F}_{LR} - Conv({\bm F}_{Ref}) \;, \\
	{\bm F}_{LR}^{'} = Deconv({\bm F}_{LR} + {\bm Res}_{LR}) \;.
	\end{array}
\right.
\end{equation} 
At last, the outputs of two branches are concatenated and passed through another convolution layer with stride 1. In this way, the details in the LR and Ref features are enhanced and aggregated, leading to more representative features.

\subsection{Loss Functions}
\label{sec:loss}

\noindent\textbf{Reconstruction loss. } We adopt $L_{1}$ loss as the reconstruction loss as
\begin{align}
\mathcal{L}_{rec} = \| {\bm I}_{HR} - {\bm I}_{SR} \|_{1} \;,
\end{align}
where ${\bm I}_{HR}$ and ${\bm I}_{SR}$ denote the ground truth image and the network output.

\noindent\textbf{Perceptual loss. } The perceptual loss is expressed as
\begin{align}
\mathcal{L}_{per} = \| {\phi}_{i} ({\bm I}_{HR}) - {\phi}_{i} ({{\bm I}_{SR}}) \|_{2} \;,
\end{align}
where ${\phi}_{i}$ denotes the $i$-th layer of VGG19. Here we use $conv5\_4$.

\noindent\textbf{Adversarial loss. } The adversarial loss~\cite{gan} $\mathcal{L}_{adv}$ is effective in generating visually pleasing images with natural details. We adopt the Relativistic GANs~\cite{rgan}:
\begin{align}
	\begin{split}
		\mathcal{L}_{D} = &-{\mathbb{E}_{{\bm I}_{HR}}} \left[ \log (D({\bm I}_{HR}, {\bm I}_{SR})) \right] \\
		&-{\mathbb{E}_{{\bm I}_{SR}}} \left[ \log (1-D({\bm I}_{SR}, {\bm I}_{HR})) \right] \;,
	\end{split}
\end{align}
\vspace{-0.25in}
\begin{align}
	\begin{split}
		\mathcal{L}_{G} = &-{\mathbb{E}_{{\bm I}_{HR}}} \left[ \log (1-D({\bm I}_{HR}, {\bm I}_{SR})) \right] \\
		&-{\mathbb{E}_{{\bm I}_{SR}}} \left[ \log (D({\bm I}_{SR}, {\bm I}_{HR})) \right] \;.
	\end{split}
\end{align}

\noindent\textbf{Full objective. } Our full objective is defined as
\begin{align}
\mathcal{L} = {\lambda}_{rec}\mathcal{L}_{rec} + {\lambda}_{per}\mathcal{L}_{per} + {\lambda}_{adv}\mathcal{L}_{adv} \;.
\end{align}

\begin{table}[t]
	\setlength{\belowcaptionskip}{0pt}
	\scalebox{0.85}{
		\begin{tabular}{l|cccc}
			\hline 
			Algorithm & CUFED5 & Sun80 & Urban100 \\ 
			\hline 
			SRCNN~\cite{srcnn} & 25.33 / 0.745 & 28.26 / 0.781 & 24.41 / 0.738 \\ 
			MDSR~\cite{edsr} & 25.93 / 0.777 & 28.52 / 0.792 & 25.51 / 0.783 \\ 
			RDN~\cite{rdn} & 25.95 / 0.769 & 29.63 / 0.806 & 25.38 / 0.768 \\ 
			RCAN~\cite{rcan} & 26.15 / 0.767 & 29.86 / 0.808 & 25.40 / 0.765 \\ 
			HAN~\cite{han} & 26.15 / 0.767 & 29.91 / 0.809 & 25.41 / 0.765 \\ 
			SRGAN~\cite{srgan} & 24.40 / 0.702 & 26.76 / 0.725 & 24.07 / 0.729 \\ 
			ENet~\cite{enet} & 24.24 / 0.695 & 26.24 / 0.702 & 23.63 / 0.711 \\ 
			ESRGAN~\cite{esrgan} & 23.84 / 0.693 & 26.77 / 0.705 & 23.25 / 0.695 \\ 
			\hline 
			CrossNet~\cite{crossnet} & 25.48 / 0.764 & 28.52 / 0.793 & 25.11 / 0.764 \\ 
			SRNTT~\cite{srntt} & 25.61 / 0.764 & 27.59 / 0.756 & 25.09 / 0.774 \\ 
			SRNTT-rec~\cite{srntt} & 26.24 / 0.784 & 28.54 / 0.793 & 25.50 / 0.783 \\ 
			TTSR~\cite{ttsr} & 25.53 / 0.765 & 28.59 / 0.774 & 24.62 / 0.747 \\ 
			TTSR-rec~\cite{ttsr} & {\color{blue}27.09 / 0.804} & {\color{blue}30.02 / 0.814} & {\color{blue}25.87 / 0.784} \\ 
			\textbf{MASA} & 24.92 / 0.729 & 27.12 / 0.708 & 23.78 / 0.712 \\
			\textbf{MASA-rec} & {\color{red} 27.54 / 0.814} & {\color{red} 30.15 / 0.815} & {\color{red} 26.09 / 0.786} \\ 
			\hline 
	\end{tabular}}
	\caption{PSNR/SSIM comparison among different SR methods on 3 testing datasets. Methods are grouped by SISR~(top) and RefSR~(bottom). The best and the second best results are colored in red and blue.}
	\label{table:quantitative}
\vspace{-0.1in}
\end{table}

\section{Experiments}
 
\subsection{Datasets}
 
Our model is trained on CUFED5~\cite{srntt} dataset with a $\times4$ upscale factor following the setting in \cite{srntt,ttsr}. CUFED5 is composed of 11,871 training pairs. Each pair contains an original HR image and a corresponding reference image at 160 $\times$ 160 resolution. To validate the generalization capacity of our model, we test it on three popular benchmarks: CUFED5 testing set, Urban100~\cite{urban100} and Sun80~\cite{sun2012super}. 

CUFED5 testing set consists of 126 testing pairs, and each HR image is accompanied by 4 reference images with different similarity levels based on SIFT~\cite{sift} feature matching. We stitch 4 references to one image, same as that of~\cite{ttsr}, during testing. Urban100 contains 100 building images without references, and we take the LR image as the reference such that the network explores self-similarity of input images. Sun80 contains 80 natural images, each paired with several references. We randomly sample one of them as the reference image. All results of PSNR and SSIM are evaluated on the Y channel of YCbCr color space.

\subsection{Implementation Details}
 
The encoder consists of 3 building blocks, each composed of 1 convolutional layer and 4 ResBlocks~\cite{resnet}. The fusion module consists of 1 spatial adaptation module, 1 dual residual aggregation module, several convolutional layers and ResBlocks. The numbers of ResBlocks in $1\times$, $2\times$ and $4\times$ fusion modules are $12$, $8$, and $4$. The number of all intermediate channels is 64. The activation function is ReLU. No batch normalization~(BN) layer is used in our network.

In MEM, the LR block size is set to $8\times 8$. The Ref$\downarrow$ block size is set to $\frac{12H_{Ref\downarrow}}{H_{LR}}\times \frac{12W_{Ref\downarrow}}{W_{LR}}$, where $H_{LR}$, $W_{LR}$ and $H_{Ref\downarrow}$, $W_{Ref\downarrow}$ are the height and width of the LR image and the Ref$\downarrow$ image, respectively. The patch size is set to $3\times 3$.
The discriminator structure is the same as that adopted in \cite{esrgan}. 
We train our model with the Adam optimizer by setting ${\beta}_{1}=0.9$ and ${\beta}_{2}=0.999$. The learning rate is set to 1e-4 and the batch size is 9. The weight coefficients ${\lambda}_{rec}$, ${\lambda}_{per}$ and ${\lambda}_{adv}$ are 1, 1 and 5e-3, respectively. 

\begin{table}
	\setlength{\belowcaptionskip}{0pt}
	\centering
	\scalebox{0.72}{
		\begin{tabular}{l|rrrr}
			\hline 
			Algorithm & FLOPS-M~(G) & FLOPS-T~(G) & Param.~(M) & Runtime~(ms) \\ 
			\hline 
			CrossNet~\cite{crossnet} & -~~~~ & {\color{red} 348.31}~~~~ & 35.18~~~~ & {\color{red} 98.7}~~~~ \\ 
			SRNTT~\cite{srntt} & 6,005.78~~~~ & 6,500.70~~~~ & 5.75~~~~ & 4,161.6~~~~ \\ 
			TTSR~\cite{ttsr} & 618.48~~~~ & 1,044.28~~~~ & 6.99~~~~ & 199.8~~~~ \\ 
			\textbf{MASA} & {\color{red} 8.84}~~~~ & {\color{blue} 367.93}~~~~ & {\color{red} 4.03}~~~~ & {\color{blue} 141.1}~~~~ \\
			\hline 
	\end{tabular}}
	\caption{FLOPS of matching steps~(FLOPS-M), total FLOPS~(FLOPS-T), number of network parameters and runtime comparisons among different RefSR methods. CrossNet~\cite{crossnet} is an alignment-based method, while the others are matching-based methods. }
	\label{table:flops}
	\vspace{-0.1in}
\end{table}

\begin{figure*}[t]
	\begin{center}
		\includegraphics[width=1.\linewidth]{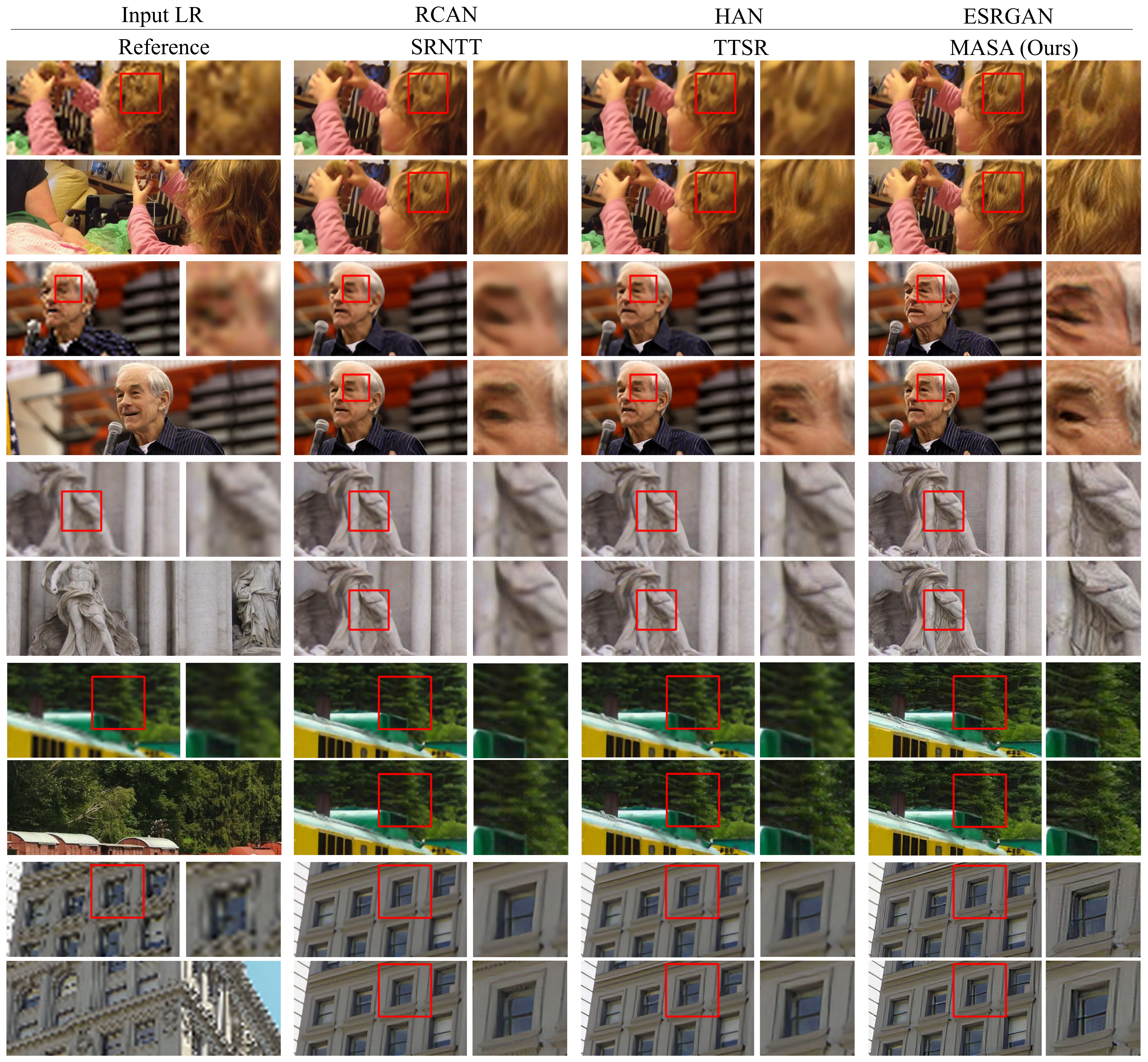}
	\end{center}
	\vspace{-0.2in}
	\caption{Visual comparison among different SR methods on the CUFED5 testing set~(top two examples), Sun80~\cite{sun2012super}~(the third and the fourth example) and Urban100~\cite{urban100}~(the last example). This figure is best viewed by zoom-in. }
	\label{fig:qualitative}
	\vspace{-0.06in}
\end{figure*}

\subsection{Comparison with State-of-the-Art Methods}
 
We compare our proposed model with previous state-of-the-art SISR and RefSR methods. SISR methods include SRCNN~\cite{srcnn}, MDSR~\cite{edsr}, RDN~\cite{rdn}, RCAN~\cite{rcan} and HAN~\cite{han}. GAN-based SISR methods include SRGAN~\cite{srgan}, ENet~\cite{enet} and ESRGAN~\cite{esrgan}. Among all these methods, RCAN and HAN achieved the best performance on PSNR, and ESRGAN is considered state-of-the-art in terms of visual quality. Some recent RefSR methods are also included, \ie, CrossNet~\cite{crossnet}, SRNTT~\cite{srntt}, TTSR~\cite{ttsr}. All the models are trained on the CUFED5 training set, and tested on the CUFED5 testing set of Sun80 and Urban100. The scale factor in all experiments is $\times4$.

\vspace{4pt}
\noindent\textbf{Quantitative evaluations. } For fair comparison with other MSE minimization based methods on PSNR and SSIM, we train another version of MASA by only minimizing the reconstruction loss, denoted as MASA-rec. 

Table~\ref{table:quantitative} shows the quantitative comparisons on PSNR and SSIM, where the best and the second best results are colored in red and blue.
As shown in Table~\ref{table:quantitative}, our model outperforms state-of-the-art methods on all three testing sets.
 
We also compare the FLOPS, the number of network parameters and runtime with other RefSR methods in Table~\ref{table:flops}, where FLOPS-M denotes the FLOPS of the matching steps, and FLOPS-T denotes the total FLOPS. The FLOPS is calculated on input of a $128\times128$ LR image and a $512\times512$ Ref image. Our model yields the smallest number of parameters and the second-best FLOPS and runtime with the best performance on PSNR/SSIM. Though the alignment-based method CrossNet~\cite{crossnet} has the smallest FLOPS and runtime, its performance on PSNR/SSIM is not on top when compared with other methods in Table~\ref{table:quantitative}. 

\vspace{4pt}
\noindent\textbf{Qualitative evaluations. } We show visual comparison between our model and other SISR and RefSR methods in Fig.~\ref{fig:qualitative}. Our proposed MASA outperforms other methods in terms of visual quality, generating more fine details without introducing many unpleasing artifacts in general. MASA produces a higher level of natural hair, wrinkle, and leaf texture.

\begin{figure*}[t]
	\centering
	\hspace{-0.04\textwidth}
	\begin{subfigure}[b]{0.3\textwidth}
		\centering
		\includegraphics[width=\textwidth]{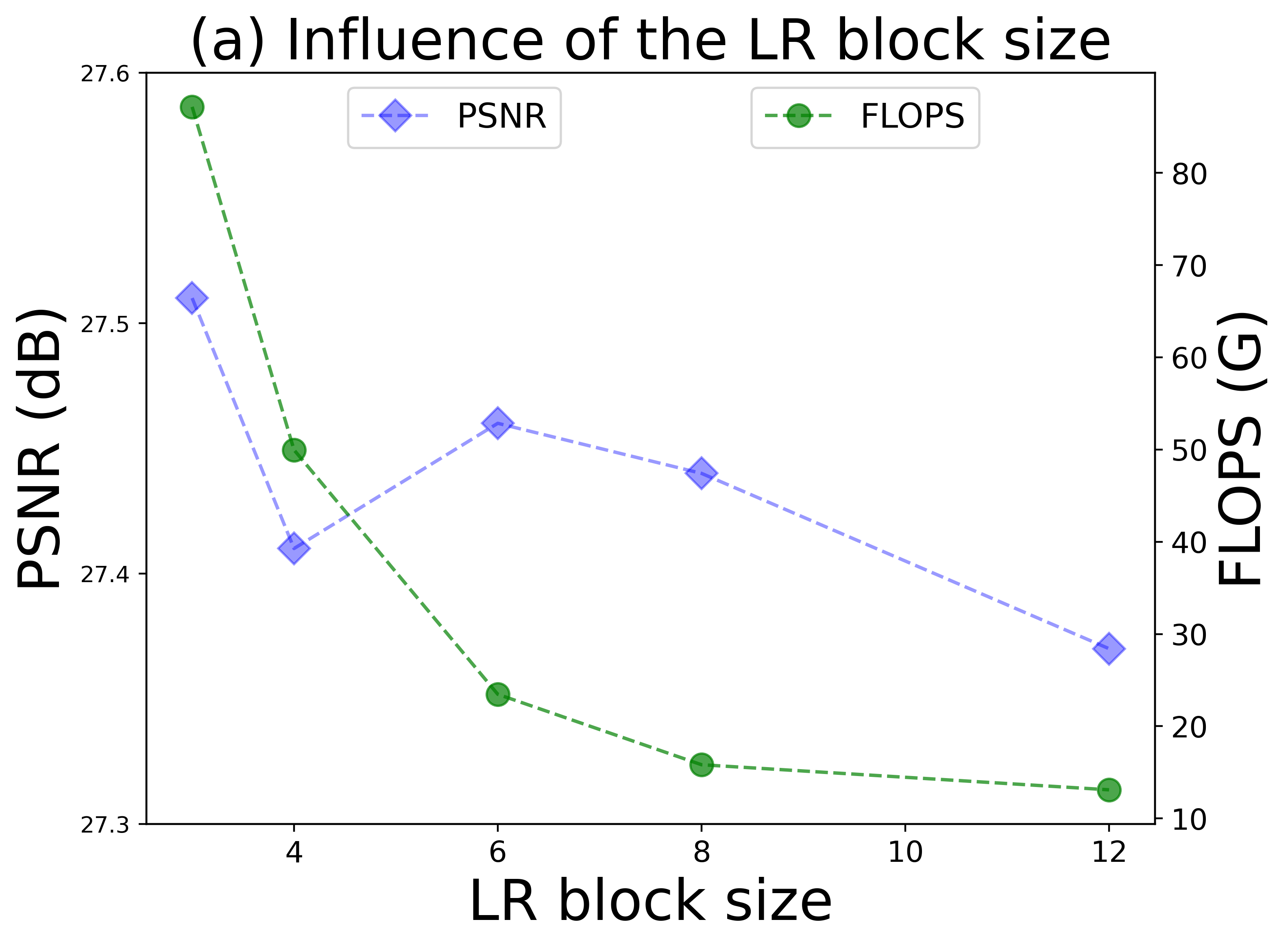}
		\label{fig:aba_size_a}
	\end{subfigure}
	\hspace{0.04\textwidth}
	\begin{subfigure}[b]{0.31\textwidth}
		\centering
		\includegraphics[width=\textwidth]{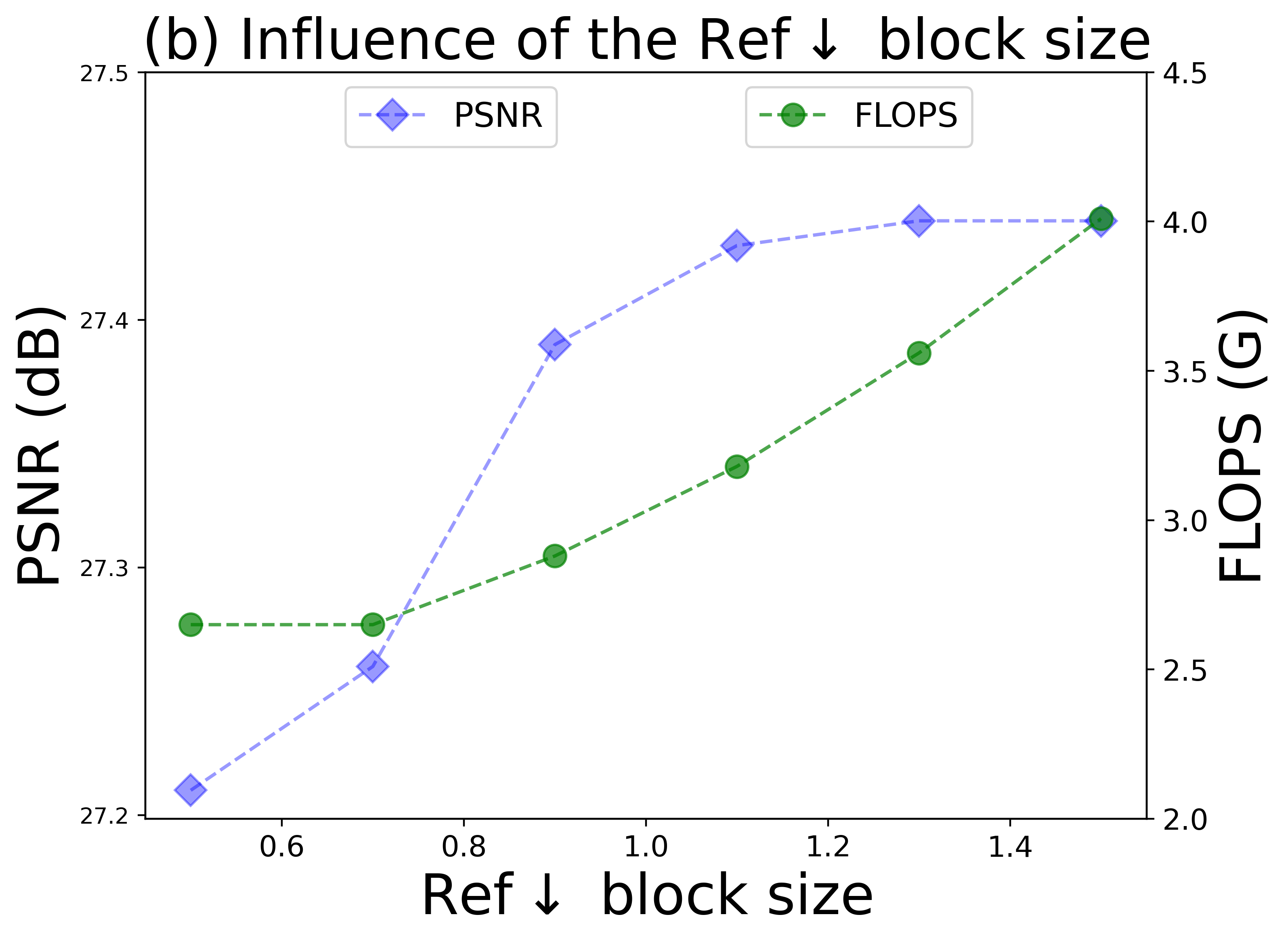}
		\label{fig:aba_size_b}
	\end{subfigure}
	\hspace{0.04\textwidth}
	\begin{subfigure}[b]{0.31\textwidth}
		\centering
		\includegraphics[width=\textwidth]{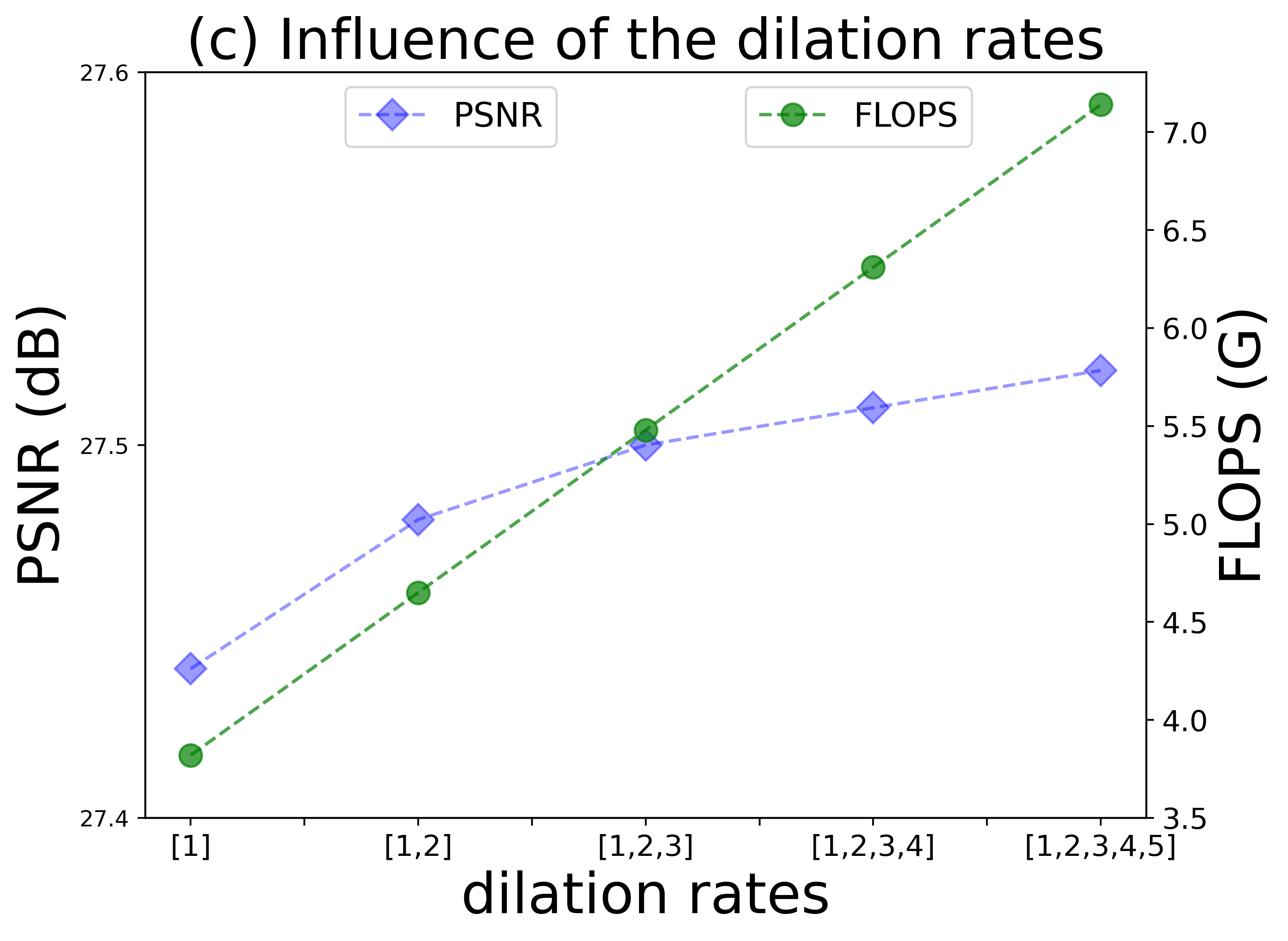}
		\label{fig:aba_size_c}
	\end{subfigure}
	\hspace{-0.04\textwidth}
	\vspace{-0.18in}
	\caption{Influence of different LR block sizes, Ref$\downarrow$ block sizes and dilation rates on PSNR and FLOPS. (a) Influence of LR block sizes. (b) Influence of Ref$\downarrow$ block sizes. (c) Influence of dilation rates.}
	\label{fig:aba_size}
	\vspace{-0.10in}
\end{figure*} 

\subsection{Ablation Study}
  
In this section, we conduct several ablation studies to investigate our proposed method. We analyze the influence of different block sizes and dilation rates used in the coarse matching stage. We also verify the effectiveness of the proposed Spatial Adaptation Module and the Dual Residual Aggregation Module. 

\noindent\textbf{Influence of block sizes and dilation rates. } In the matching~\&~extraction module,
the LR block size, the Ref$\downarrow$ block size and the dilation rates are key factors to balance the matching accuracy and efficiency. Thus we analyze the influence of these three hyper-parameters on the CUFED5 testing set. Fig.~\ref{fig:aba_size} shows the ablation results. We only show the FLOPS of batch matrix-matrix product operations.

Fig.~\ref{fig:aba_size}(a) shows the influence of the LR block size. 
It can be seen that as the LR block size increases, PSNR and FLOPS both drop, indicating the decreasing matching accuracy and computational cost. We also test the case that the size of LR block is $1\times 1$, the PSNR reaches 27.60 dB while the FLOPS sharply goes up to 787.77G, which is not shown in Fig.~\ref{fig:aba_size}(a). 

Fig.~\ref{fig:aba_size}(b) shows the influence of the Ref$\downarrow$ block size.
The PSNR and FLOPS increase as the increasing of the Ref$\downarrow$ block size. Because larger Ref$\downarrow$ block size boosts the matching accuracy in the fine matching stage. However, when the Ref$\downarrow$ block size increases to some extent, the growth of PSNR slows down. On the other hand, since patch matching has to be performed on larger blocks, the computational cost increases inevitably.

As illustrated in Fig.~\ref{fig:aba_size}(c), the more combinations of different dilation rates exist, the higher PSNR can be obtained. Since larger dilation rates cover a larger area in the LR block, it leads to more accurate coarse matching.

\begin{figure}
	\begin{center}
		\includegraphics[width=0.99\linewidth]{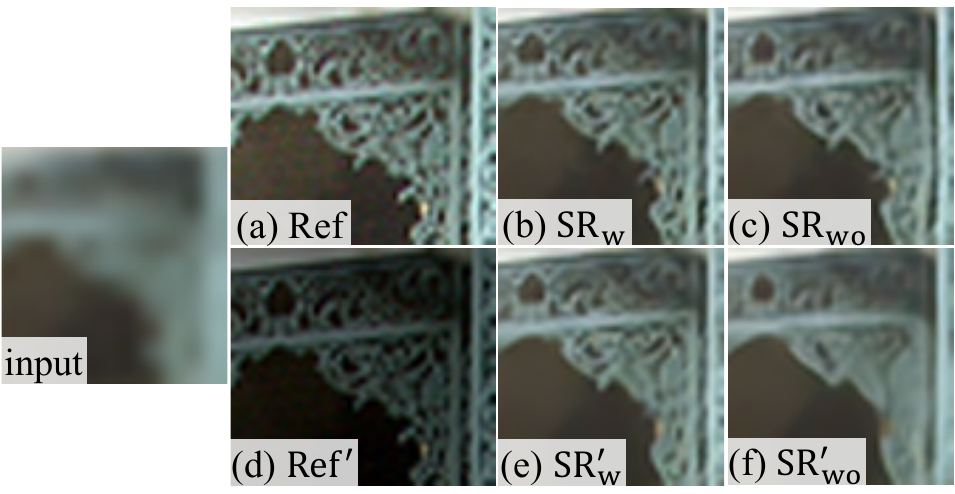}
	\end{center}
	\vspace{-0.15in}
	\caption{Ablation study on the spatial adaptation module. The results with the subscript 'w' are the outputs of the model with SAM, while those with the subscript 'wo' are the outputs of the baseline model without SAM. }
	\label{fig:aba_pa}
\vspace{-0.1in}
\end{figure}

\vspace{4pt}
\noindent\textbf{Effect of spatial adaptation module. } The spatial adaptation module plays the role of aligning the distribution of the Ref features to that of the LR features. 
{As shown in Table~\ref{table:aba_pa}, compared with the baseline~(without any normalization), SAM improves the PSNR and SSIM by 0.22 dB and 0.007, \textit{only} introducing 0.15M more parameters. We also compare SAM with other normalization methods, including adaptive instance norm.~(AdaIN)~\cite{huang2017arbitrary} and SPADE~\cite{spade}. We found that SPADE is not suitable for this task, and the performance of AdaIN is similar with that of the baseline.
	
Qualitative comparison is visualized in Fig.~\ref{fig:aba_pa}. We first test the model with SAM and the baseline model on the Ref image, which has a similar luminance with the LR image. Then we change the luminance of the Ref image~(denoted by $\rm Ref'$), and test two models on it. As shown in Fig.~\ref{fig:aba_pa}, the model with SAM performs better in both cases. Besides, when testing on $\rm Ref'$ with changing luminance, the performance of the model with SAM almost keeps the same, while the performance of baseline drops significantly. This demonstrates that SAM is robust in handling different Ref images.}

\begin{table}
	\setlength{\belowcaptionskip}{0pt}
	\centering
	\footnotesize
	\scalebox{0.85}{
		\begin{tabular}{l|cccc}
			\hline 
			 & baseline & AdaIN & SPADE & SAM~(ours) \\ 
			\hline 
			PSNR / SSIM & 27.32 / 0.807 & 27.30 / 0.806 & 24.46 / 0.688 & 27.54 / 0.814 \\ 
			param. & 3.88M & 3.88M & 3.99M & 4.03M \\ 
			\hline 
	\end{tabular}}
	\caption{Ablation study on the spatial adaptation module. }
	\label{table:aba_pa}
\vspace{-0.03in}
\end{table}

\vspace{4pt}
\noindent\textbf{Effect of dual residual aggregation module. } To verify the effectiveness of the dual residual aggregation module (DRAM), we conduct ablation study on 4 models as shown in Table~\ref{table:aba_DRAM}. Model 1 simply concatenates the LR features with the Ref features and feeds them into a convolution layer. Model 2 only keeps the LR branch of the DRAM, and Model 3 only keeps the Ref branch. Model 4 is the proposed DRAM. In Table~\ref{table:aba_DRAM}, it is clear that DRAM outperforms Model 1 by 0.11 dB.

\begin{table}
	\setlength{\belowcaptionskip}{0pt}
	\centering
	\scalebox{0.7}{
		\begin{tabular}{l|cccc}
			\hline 
			Model & Model 1 & Model 2 & Model 3 & Model 4 \\ 
			\hline 
			Ref branch & \xmark & \xmark & \cmark & \cmark \\ 
			LR branch & \xmark & \cmark & \xmark & \cmark \\ 
			\hline
			PSNR / SSIM & 27.43 / 0.809 & 27.51 / 0.813 & 27.48 / 0.811 & 27.54 / 0.814 \\
			param. & 3.73M & 3.88M & 3.88M & 4.03M \\
			\hline 
	\end{tabular}}
	\caption{Ablation study on dual residual aggregation. }
	\label{table:aba_DRAM}
\vspace{-0.10in}
\end{table}

\section{Conclusion}

In this paper, we have proposed MASA-SR, a new end-to-end trainable network for RefSR. It features a coarse-to-fine correspondence matching scheme to reduce the computational cost significantly, while achieving strong matching and transfer capability. Further, a novel Spatial Adaptation Module is designed to boost the robustness of the network when dealing with Ref images with different distributions. Our method achieves state-of-the-art results both quantitatively and qualitatively across different datasets.


{\small
\bibliographystyle{ieee_fullname}
\bibliography{egbib}
}


\newpage

\appendix

\section{Appendix}

\subsection{Details of Ablation Study}

In the ablation study of block sizes and dilation rates, it should be noted that if the size of the LR block is $m\times n$, then the basic size of the Ref$\downarrow$ block is set to $\frac{mH_{Ref\downarrow}}{H_{LR}}\times \frac{nW_{Ref\downarrow}}{W_{LR}}$, where $H_{Ref\downarrow}$, $W_{Ref\downarrow}$ and $H_{LR}$, $W_{LR}$ are the height and width of the Ref$\downarrow$ feature and the LR feature, respectively. We multiply the basic size of different scale factors in the ablation study, and {\textit{use the scale factors to denote the Ref$\downarrow$ block size}} in Fig.~{\color{red}{5}}(b) of body text and in the following descriptions.

\noindent\textbf{Influence of LR block sizes.} The FLOPS in Fig.~{\color{red}{5}}(a) is computed by taking as input a $192\times 192$ LR image and a $768\times 768$ Ref image. The Ref$\downarrow$ block size is $1.5$ and the dilation is $1$.

\noindent\textbf{Influence of Ref$\downarrow$ block sizes.} The FLOPS in Fig.~{\color{red}{5}}(b) is computed on a $128\times 128$ LR image and a $512\times 512$ Ref image. The LR block size is $8$ and the dilation is $1$.

\noindent\textbf{Influence of dilation rates.} The FLOPS in Fig.~{\color{red}{5}}(c) is computed on a $120\times 120$ LR image and a $480\times 480$ Ref image. The LR block size is $12$ and the Ref$\downarrow$ block size is $1.5$.

\subsection{More Visual Results}

\begin{figure*}[t]
	\begin{center}
		\includegraphics[width=1.0\linewidth]{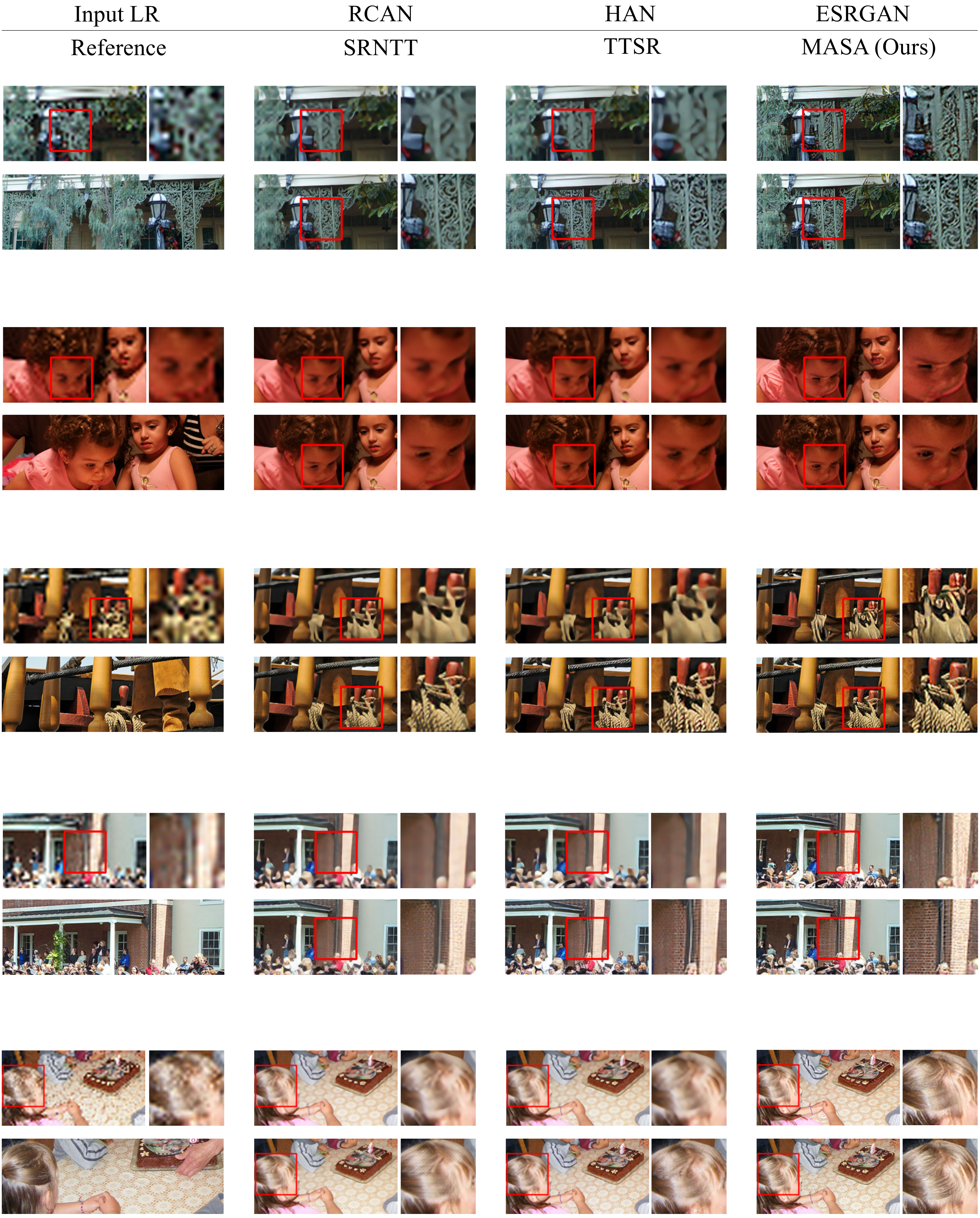}
	\end{center}
	\caption{Visual comparison among different SR methods on the CUFED5~\cite{srntt} testing set. } 
	\label{fig:comparison0}
\end{figure*}

\begin{figure*}[t]
	\begin{center}
		\includegraphics[width=1.0\linewidth]{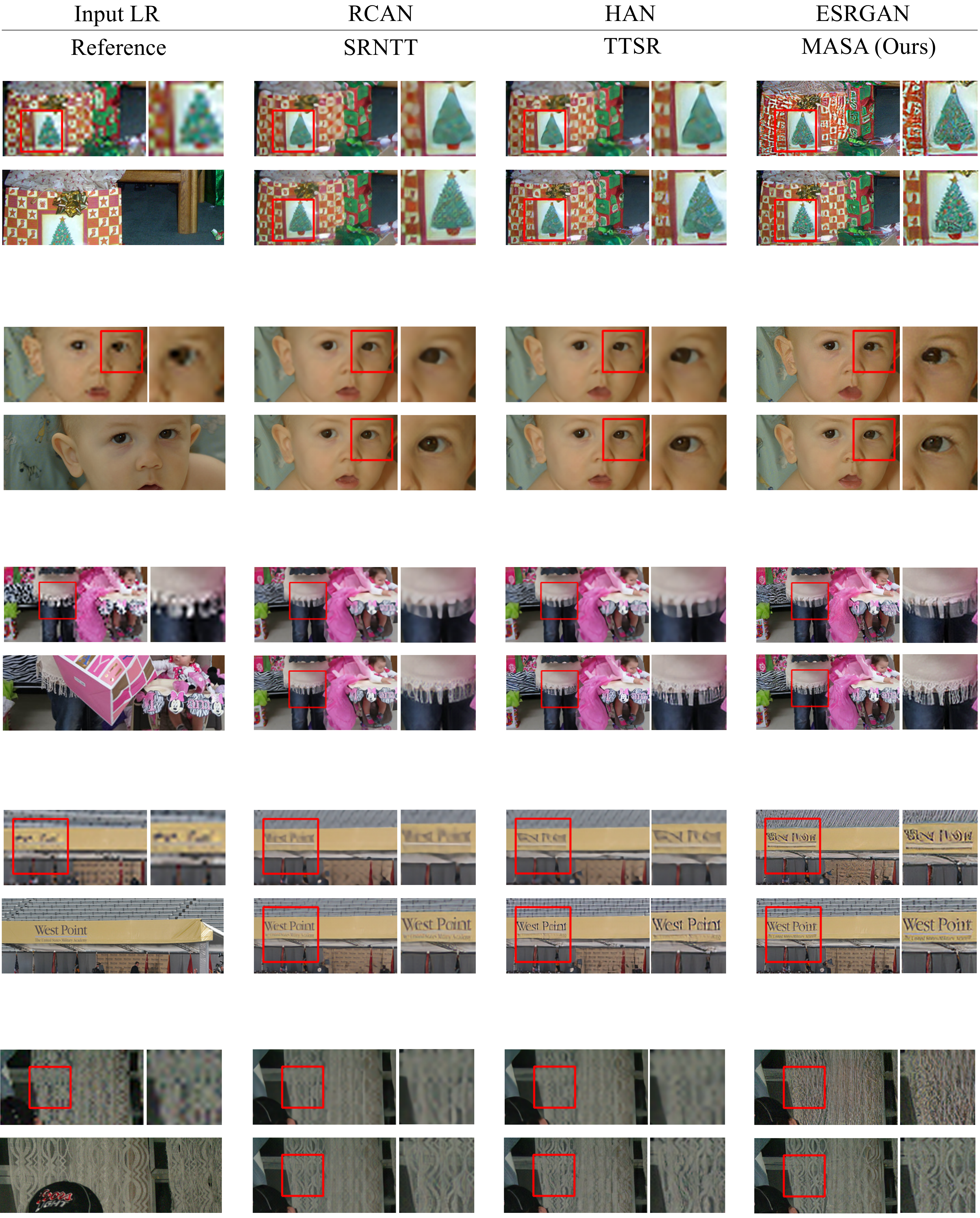}
	\end{center}
	\caption{Visual comparison among different SR methods on the CUFED5~\cite{srntt} testing set. } 
	\label{fig:comparison1}
\end{figure*}

\begin{figure*}[t]
	\begin{center}
		\includegraphics[width=1.0\linewidth]{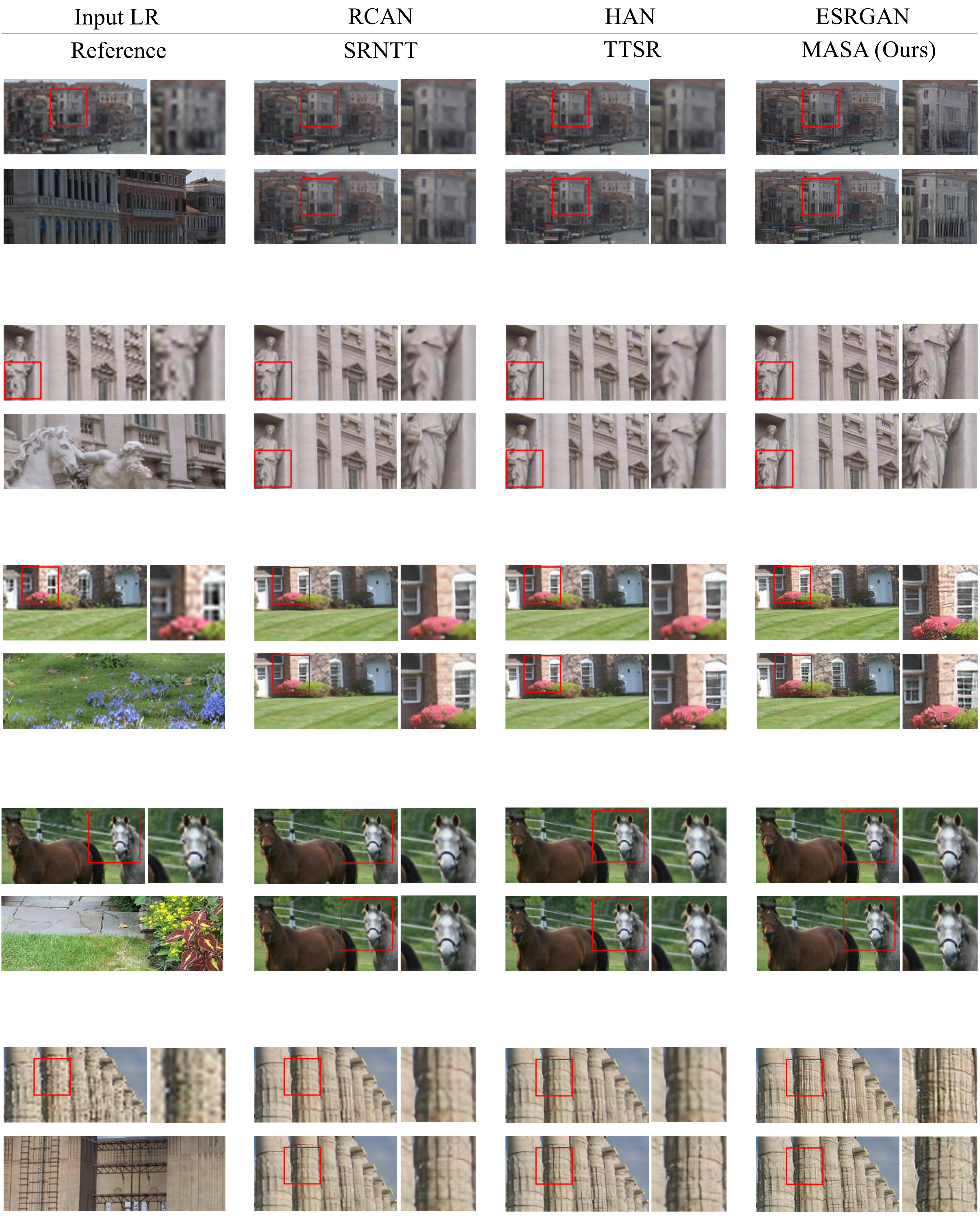}
	\end{center}
	\caption{Visual comparison among different SR methods on the Sun80~\cite{sun2012super} dataset. } 
	\label{fig:comparison2}
\end{figure*}

\begin{figure*}[t]
	\begin{center}
		\includegraphics[width=1.0\linewidth]{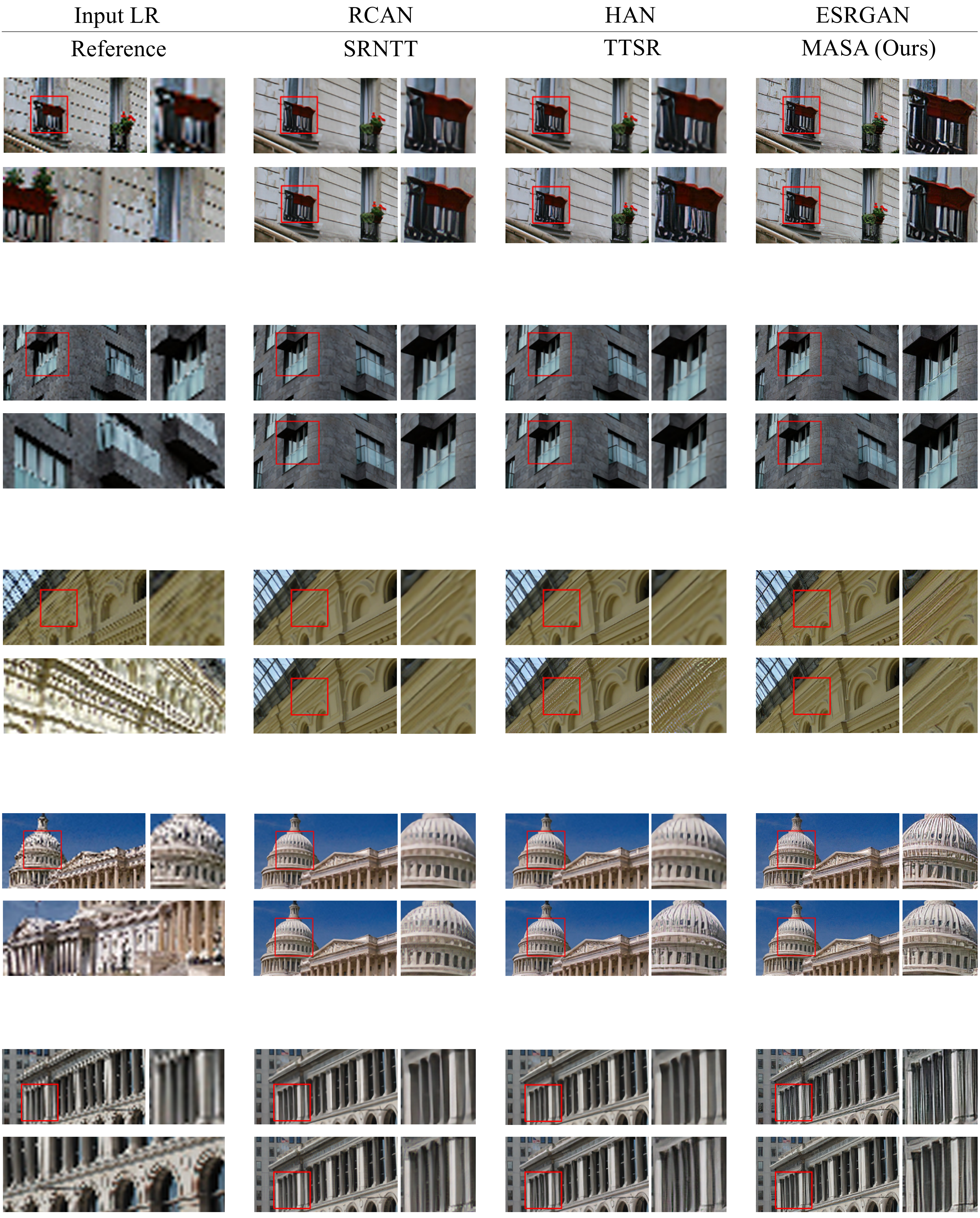}
	\end{center}
	\caption{Visual comparison among different SR methods on the Urban100~\cite{urban100} dataset. } 
	\label{fig:comparison3}
\end{figure*}

We show more visual results of the proposed MASA and other state-of-the-art methods, including RCAN~\cite{rcan}, HAN~\cite{han}, ESRGAN~\cite{esrgan}, SRNTT~\cite{srntt} and TTSR~\cite{ttsr}.
RCAN and HAN are SISR methods that have achieved the best performance on PSNR, and ESRGAN is a GAN-based SISR method that is considered state-of-the-art in visual quality. SRNTT~\cite{srntt} and TTSR~\cite{ttsr} are state-of-the-art RefSR methods. The visual comparison on CUFED5~\cite{srntt} testing set are shown in Fig.~\ref{fig:comparison0} and Fig.~\ref{fig:comparison1}, and the comparison on Sun80~\cite{sun2012super} and Urban100~\cite{urban100} are shown in Fig.~\ref{fig:comparison2} and Fig.~\ref{fig:comparison3}, respecitively.

It can be observed that our MASA can restore more regular structures and generate photo-realistic details.

\end{document}